\documentclass[letterpaper, 10 pt, conference]{ieeeconf}  % Comment this line out if you need a4paper

\IEEEoverridecommandlockouts                              % This command is only needed if 
\usepackage{graphicx} % for pdf, bitmapped graphics files
\usepackage{amsmath} % assumes amsmath package installed
\usepackage{amssymb}  % assumes amsmath package installed
\PassOptionsToPackage{dvipsnames}{xcolor}
\usepackage[inkscapelatex=false]{svg}

\usepackage{cite}
\usepackage[caption=false]{subfig}
\usepackage{tabularx}
\usepackage{wasysym}
\usepackage{booktabs}
\usepackage{soul}
\usepackage[colorlinks=false,urlcolor=blue]{hyperref}
\usepackage{makecell}
\usepackage[table]{xcolor} % in preamble
\definecolor{row_hightlight}{RGB}{235,235,255} % subtle blue-ish
\begin{document}
\title{\LARGE \bf Tensegrity Robot Endcap-Ground Contact Estimation with \\ Symmetry-aware Heterogeneous Graph Neural Network}

\author{
% Anonymous Authors
Wenzhe Tong$^*$, Yicheng Jiang$^*$, Chi Zhang, Maani Ghaffari, Xiaonan Huang
\thanks{$^*$ denotes equal contribution.}
\thanks{This work was supported by the startup fund from the Robotics Department at the University of Michigan.}
\thanks{The authors are with the Robotics Department, University of Michigan, Ann
Arbor, MI 48109 USA (e-mail: wenzhet@umich.edu; valeska@umich.edu;
zhc@umich.edu; maanigj@umich.edu; xiaonanh@umich.edu).}
}

\maketitle

\begin{abstract}
Tensegrity robots possess lightweight and resilient structures but present significant challenges for state estimation due to compliant and distributed ground contacts. This paper introduces a symmetry-aware heterogeneous graph neural network (Sym-HGNN) that infers contact states directly from proprioceptive measurements, including IMU and cable-length histories, without dedicated contact sensors. The network incorporates the robot's dihedral symmetry $D_3$ into the message-passing process to enhance sample efficiency and generalization. The predicted contacts are integrated into a state-of-the-art contact-aided invariant extended Kalman filter (InEKF) for improved pose estimation. Simulation results demonstrate that the proposed method achieves up to 15\% higher accuracy and 5\% higher F1-score using only 20\% of the training data compared to the CNN and MI-HGNN baselines, while maintaining low-drift and physically consistent state estimation results comparable to ground truth contacts. This work highlights the potential of fully proprioceptive sensing for accurate and robust state estimation in tensegrity robots. Code available at: \href{https://github.com/Jonathan-Twz/Tensegrity-Sym-HGNN}{https://github.com/Jonathan-Twz/Tensegrity-Sym-HGNN}
\end{abstract}

\section{Introduction}
    Tensegrity robots are lightweight and resilient structures capable of absorbing high impacts and traversing uneven terrain. These properties make them promising platforms for planetary exploration, disaster response, and field applications where robustness is critical. Realizing this potential requires accurate state estimation to enable real-time control, data collection, and other perceptual tasks such as localization. While motor encoders directly measure actuation, recovering the robot’s pose, velocity, and deformation requires additional sensing and principled data fusion.

Vision-based methods provide global localization but are prone to failure in degraded conditions such as dust, fog, or low illumination. Proprioceptive state estimators are agnostic to such conditions and can operate at high update rates, providing odometry for localization and feedback for control. However, proprioceptive estimation in tensegrity robots is uniquely challenging. Unlike legged robots, which benefit from structured kinematics and well-defined foot contacts, tensegrity robots exhibit distributed and dynamic contact events. Contact can occur at any node or rod, and embedding dedicated contact sensors is not always practical. Inferring contact information reliably from proprioception is therefore a key challenge and an enabling capability.

In this work, inspired by the fact that tensegrity robots are mechanically designed in the form of graphs, we propose a symmetry-aware contact estimation framework based on the symmetry-aware heterogeneous graph neural network (Sym-HGNN). The tensegrity morphology is naturally represented as a graph, with vertices corresponding to rod endcaps and edges representing cables and rods. Importantly, the structure exhibits dihedral symmetry $D_3$, which we encode into the GNN architecture. By enforcing equivariance under these symmetry transformations, the network produces consistent contact predictions across equivalent configurations and movements and improves generalization over the baseline CNN and MI-HGNN\cite{lin2021legged, butterfield2025mi}. 
We further integrate the estimated contacts into a contact-aided Invariant Extended Kalman Filter (InEKF) for state estimation. 
% The estimator fuses IMU and actuator data with geometric constraints (rod lengths and chirality) and incorporates learned contact events as observation updates.
% We validate our method through simulation and \twz{hardware experiments} on a 3-bar tensegrity robot, showing that the proposed contact estimation method can be used for contact-aware estimation and provides satisfactory results.

The contributions of this paper are as follows:
\begin{enumerate}
    \item A Sym-HGNN framework for contact estimation in tensegrity robots that requires no dedicated contact sensors;
    \item Integration of learned contact information into a contact-aided InEKF for real-time state estimation;
    \item Experimental validation in simulation demonstrating accurate and robust estimation.
    % \item Code will be open-sourced after receiving the final decision.
\end{enumerate}

\begin{figure}
    \centering
    \includegraphics[width=\linewidth]{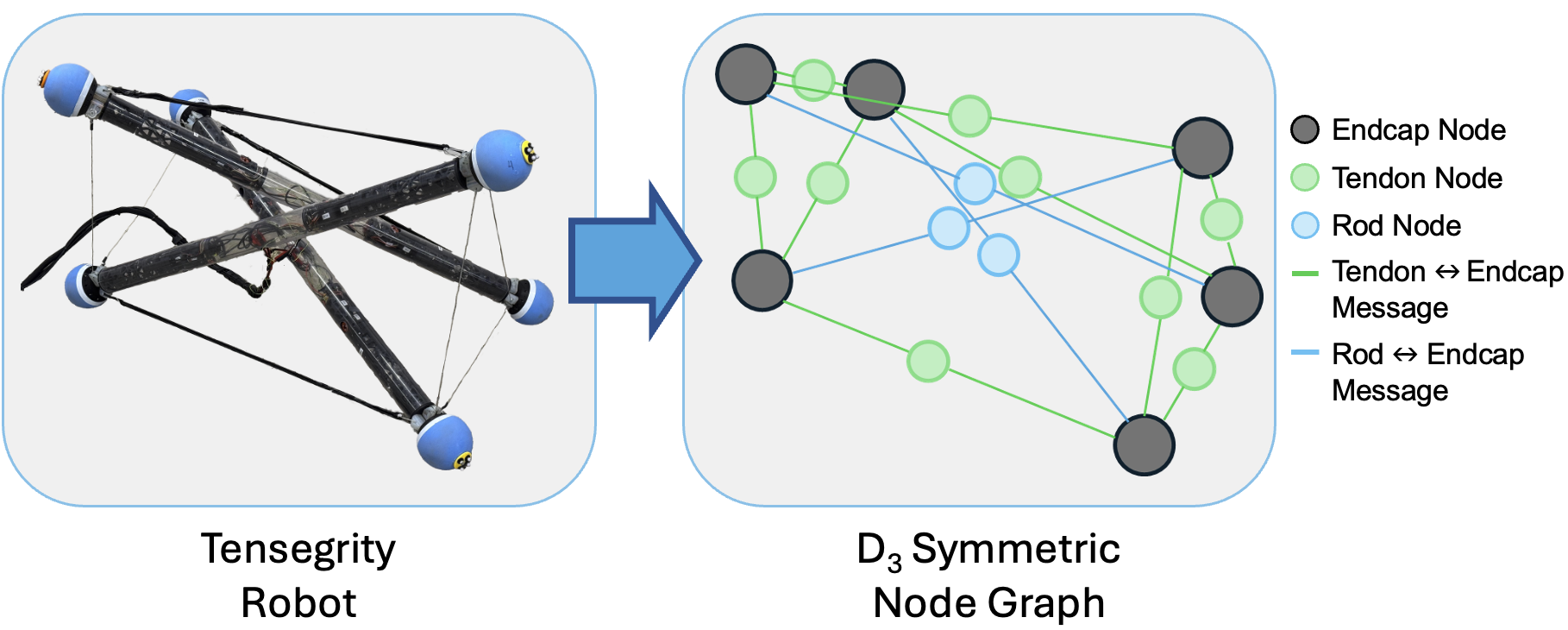}
    \caption{The proposed symmetry-aware heterogeneous graph neural network (Sym-HGNN) represents a tensegrity robot as a $D_3$-symmetric node graph, where rods, tendons, and endcaps form typed connections for structure-aware message passing and ground-contact inference.}
    \label{fig:graph-construction}
\end{figure}

\section{Related Work}
    %  tensegrity robot contact sensors
Early tensegrity platforms established onboard sensing capabilities but seldom focused explicitly on contact, relying primarily on proprioception. SUPERball \cite{caluwaerts2016state, sabelhaus2014hardware, sabelhaus2015system, vespignani2018design} integrated IMUs in strut endcaps, cable-length/torque sensing, and external ultra-wideband ranging; its UKF fused IMU orientation and cable rest lengths to estimate shape and pose during rolling, demonstrating that cable/IMU signals carry rich state and contact cues even without dedicated foot sensors. A 6-bar tensegrity robot wrapped with robotic skins \cite{booth2021surface} showed that surface strain sensing can reconstruct node locations and orientations. Dedicated electrical-resistance sensors mounted on the endcaps provide contact information and further support contact-aware locomotion.

% tensegrity robot state estimations
More recent proprioceptive estimators reduce reliance on exteroception by using stretch-tendon sensors that directly measure inter-endcap distances \cite{johnson2022sensor}, and by combining IMUs and motor encoders with geometric constraints \cite{tong2025tensegrity}. These signals are natural inputs for contact inference when the shape deviates from the free-space model. Overall, the literature establishes the main sensing modalities (IMUs, motor encoders, strain sensors) and filters (UKF/InEKF), but leaves a gap in estimating explicit contact states from purely proprioceptive signals on tensegrity robots.

% contact sensing by other robots
Other contact-aware robots, such as legged systems, also utilize foot-force or pressure sensors when available \cite{ruppert2020foottile, zhang2021tactile, wang2024research}. More recently, contacts are inferred from proprioception and models—for example, momentum-residual observers and probabilistic estimators that fuse IMU, kinematics, and dynamics. \cite{hwangbo2016probabilistic} used HMM-style fusion for foot contact without foot sensors; \cite{camurri2017probabilistic} presented probabilistic contact and impact detection for ANYmal; and \cite{bledt2018contact} proposed an EKF that fuses leg-force estimates with kinematic and phase cues on Cheetah 3. 

% Learning based contact sensing
Learning-based approaches have improved robustness to terrain and noise and enabled anticipatory contact cues in legged systems (e.g., online learning that reduces touchdown latency), but most work still treats each leg independently or with flat feature encoders. 
Lin et al. \cite{lin2021legged} trained a CNN-based proprioceptive contact estimator using only onboard signals and predicted quadruped foot-contact events across terrains. The learned contact events were then injected as constraints into an Invariant Extended Kalman Filter (InEKF) \cite{hartley2020contact} to improve legged odometry. MI-HGNN \cite{xie2024morphological} builds a morphology-informed heterogeneous GNN over robot joints and links to predict legged contact states with better generalization and parameter efficiency. 
% Graph based learning methods
% add MS-HGNN
% , while tactile GNNs leverage spatial relations among taxels for contact and pose inference.

% Relationship to this work
These works motivate a tensegrity-native GNN that treats endcaps as nodes and cables or rods as edges, fusing IMU features with cable length/tension features to infer which nodes are in contact. Compared to per-sensor thresholds or monolithic CNN/MLP encoders, a graph formulation encodes structural coupling and supports inductive generalization across robot morphologies and gaits.

% It addresses the distributed, multi-point contact patterns unique to tensegrity robots. 

\section{Preliminaries}
    \subsection{3-bar Tensegrity Robot and $D_3$ Symmetry}
    % Definition of robots
    % Hardware introductions
    % Tensegrity robot state estimation methods
    
    \begin{figure}[htbp]
        \centering
        \includegraphics[width=0.99\linewidth]{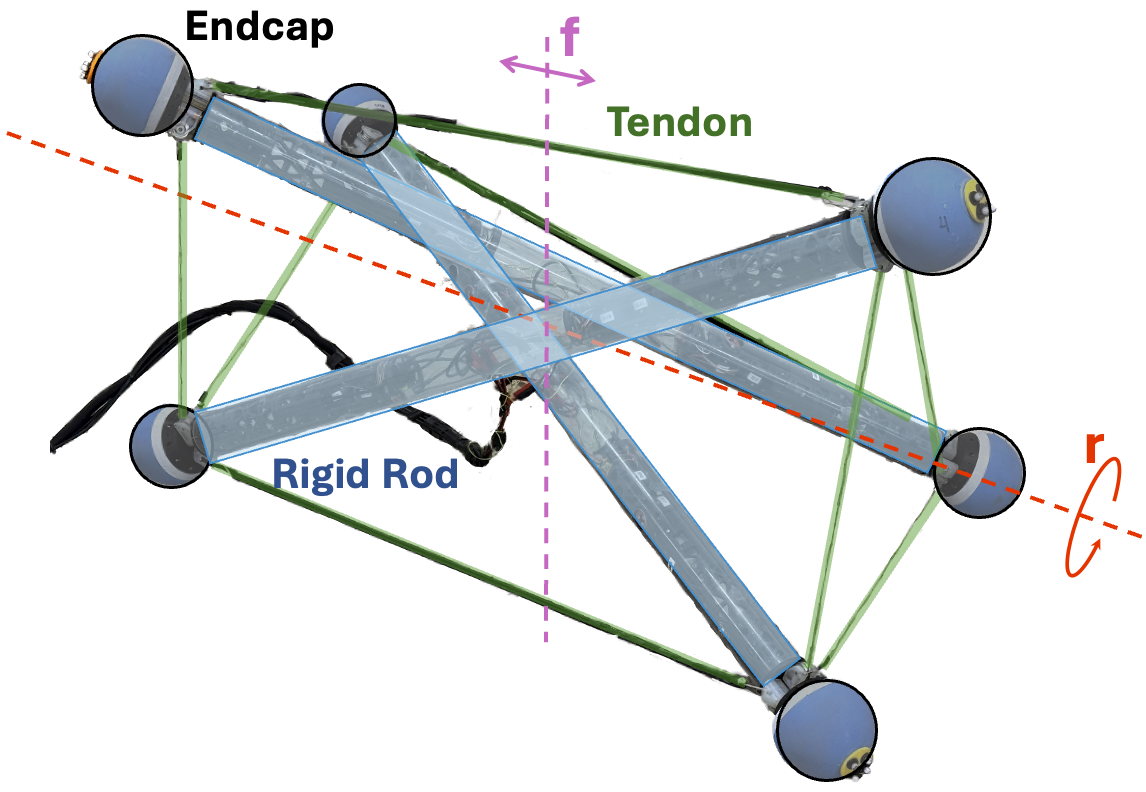}
        \caption{Structural composition and symmetry properties of the 3-bar tensegrity robot. The structure consists of rigid rods, tensile tendons, and six endcaps, exhibiting 120° rotational symmetry about longitudianl axis r and reflection symmetry across vertical plane f.}
        \label{fig:hardware}
    \end{figure}
    
    Tensegrity robots comprise rigid bars linked by tensile tendons so that bars carry compression and tendons carry tension. Locomotion is produced by modulating tendon lengths to reshape the structure and shift the center of mass (CoM) relative to active ground contacts. In a three–bar prism with six endcaps, a rolling step typically preloads tendons, tilts the body until the CoM crosses the support boundary, and pivots about one or two endcaps; persistent asymmetry in commands yields turning. 
    Geometrically, the three-bar tensegrity exhibits dihedral $D_3$ symmetry, characterized by 120° rotational symmetry about the central longitudinal axis (r) and reflection symmetry across the vertical plane (f) as shown in Fig.~\ref{fig:hardware}.

\subsection{Hardware and Onboard Sensors}
    The robot platform is a cable-actuated three-bar tensegrity\cite{mi2025design} consisting of six endcaps and nine motor-driven tendons arranged along the side cables as shown in Fig.~\ref{fig:hardware}. Each tendon is actuated by a quasi-direct drive motor with a high-resolution encoder that provides precise, high-rate length measurements $l_i$ and enables variable stiffness control. Three inertial measurement units (IMUs) are rigidly mounted on the rods, measuring linear acceleration $a_i$ and angular velocity $\omega_i$ of each rod to capture the system’s dynamic motion. All sensors and actuators interface through a Jetson Orin Nano controller, which communicates with distributed motor drivers and IMUs at low latency and high frequency.

\subsection{State Estimation for Tensegrity Robots}
    We adopt a proprioceptive state estimation pipeline\cite{tong2025tensegrity} with two stages: (i) shape reconstruction using measured cable lengths and IMU readings, which provides the robot's instantaneous shape and contact kinematics; and (ii) a contact–aided invariant Kalman filter that fuses IMU propagation with contact-based kinematic corrections to estimate the global pose. 
    
    In this framework, stable and reliable contact information is crucial — the filter assumes that detected contacts are both temporally consistent and physically valid. However, direct contact sensing in tensegrity robots is challenging: integrating dedicated contact sensors into all endcaps complicates mechanical design, increases wiring complexity, and interferes with cable routing. As a result, existing approaches relying on thresholding or geometric heuristics often yield intermittent or noisy contact detections, which degrade estimation accuracy. 
    % TODO: add citation here about previous state estimation works, j
    
    To address this limitation, we introduce a learning-based contact estimation module that infers contact states directly from proprioceptive signals. By providing temporally consistent and robust contact predictions, this component ensures reliable contact triggers for the invariant filter and significantly improves overall state estimation stability.

\section{Methodology}
    \subsection{System Overview}
    We present a real-time ground contact estimation module based on a heterogeneous graph neural network (HGNN). The proposed method utilizes temporal data from three rod-mounted IMUs (linear acceleration $a$ and angular velocity $\omega$) and nine tendon lengths $l$ to predict the contact pattern of six endcaps. We define the contact vector as
    \begin{equation}
        \mathcal{C} = [c_0, c_1, c_2, c_3, c_4, c_5], \quad c_i \in \{0,1\},
    \end{equation}
    where $c_i = 1$ denotes ground contact and $c_i = 0$ denotes no contact.  

\subsection{Graph Construction}
    \begin{figure*}
        \centering
        \includegraphics[width=\linewidth]{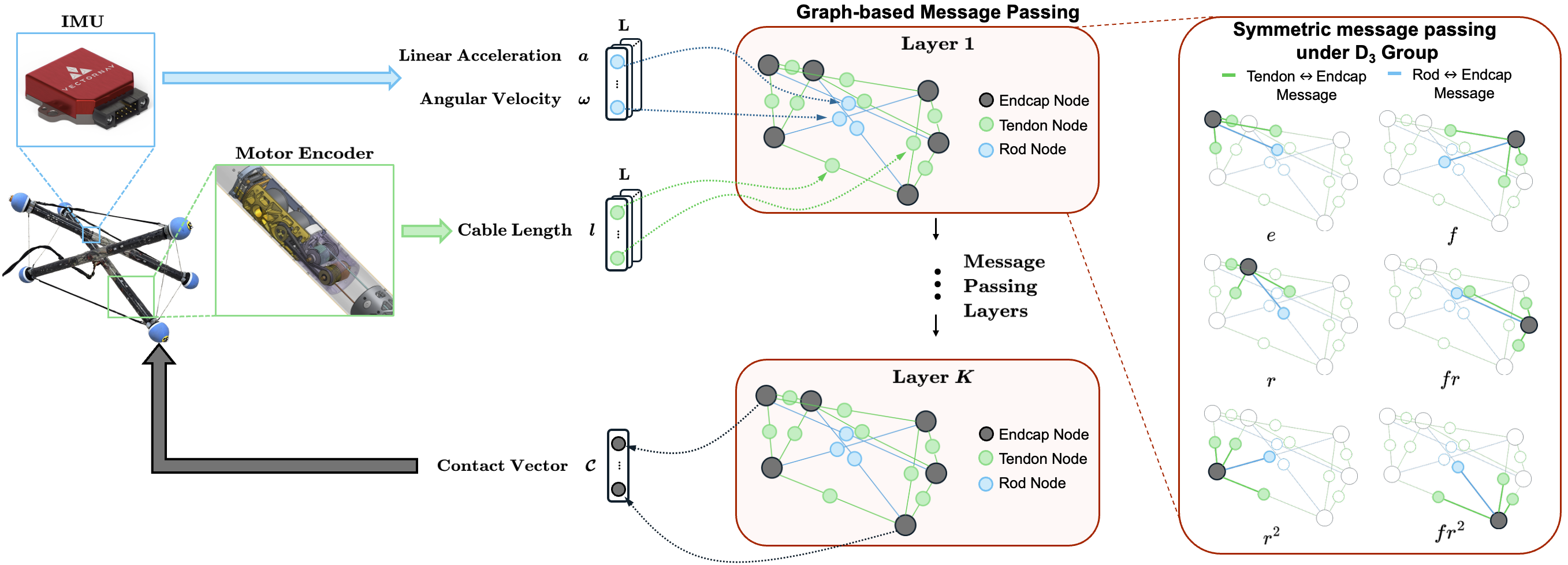}
        \caption{Overview of the proposed Sym-HGNN contact prediction pipeline. IMU and motor encoders provide proprioceptive inputs, which are propagated through a heterogeneous graph with typed edges. Message passing is performed across layers under six $D_3$-equivariant transformations $\{e, r, r^2, f, fr, fr^2\}$, enabling symmetry-aware contact inference.
        % \mgj{Add a detailed caption and also explain nodes and edges and legend for the figures.}
        }
        \label{fig:framework}
    \end{figure*}
    We represent the tensegrity morphology as a heterogeneous graph
    \[
    \mathcal{G} = (\mathcal{V}, \mathcal{E}),
    \]
    where the vertex set
    \(\mathcal{V} = \mathcal{V}_{r} \cup \mathcal{V}_{t} \cup \mathcal{V}_{e}\)
    comprises three node types: \textit{rods}, \textit{tendons}, \textit{endcaps}.
    Rod nodes encode rigid-body motion from IMU readings, tendon nodes represent actuator states via length variations, and endcap nodes denote potential contact points to be inferred.
    Edges in \(\mathcal{E}\) describe the coupling among these subsystems,
    forming a mixed topology of rigid (rod) and elastic (tendon) relations.

    \subsubsection{Data Preprocessing}
        Synchronized IMU and tendon length measurements serve as inputs for contact estimation. A fixed-length sliding window of size $L=100$ captures short-term temporal information. Within each window, a heterogeneous graph is instantiated with three rod nodes, six endcap nodes, and nine tendon nodes as shown in Fig.~\ref{fig:graph-construction}. Node features are defined as follows:
        \begin{itemize}
            \item \textbf{Rod Node}: Stacked IMU measurements, including linear acceleration $a$ and angular velocity $\omega$, over $L$ timestamps.
            \item \textbf{Tendon Node}: Tendon-length histories of dimension $L \times 1$, representing actuator states. 
            \item \textbf{Endcap Node}: Zero-initialized placeholders whose contact states are predicted by the network.
        \end{itemize}
        Per-channel z-score normalization is applied independently within each window. Directed, typed edges connect (rod$\leftrightarrow$endcap) and (tendon$\leftrightarrow$endcap), capturing structural coupling and kinematic information.
    
    \subsubsection{Physical intuition}
        The heterogeneous graph design reflects the asymmetric roles of rods and tendons in tensegrity dynamics: rods sustain compression while tendons regulate geometry through controlled tension. By conditioning message passing on edge type, the network distinguishes geometric constraints from actuation effects, capturing how motion and tension propagate through the structure.
    
    \subsubsection{Morphological symmetry}
        The three-bar tensegrity robot with six-endcap configuration exhibits the dihedral symmetry group \(D_{3}\), generated by a \(120^\circ\) rotation \(r\) and a reflection \(f\) about the vertical axis. Each symmetry operation \(g \in D_{3} = \{e, r, r^{2}, f, fr, fr^{2}\}\), as shown in Fig.~\ref{fig:framework}, induces deterministic permutations of node indices within \(\mathcal{V}_{r}\), \(\mathcal{V}_{t}\), and \(\mathcal{V}_{e}\). Embedding this group structure into the graph model ensures that geometrically equivalent configurations correspond to isomorphic subgraphs, thereby enabling weight sharing across symmetry-related instances and enhancing data efficiency and generalization.

    % \begin{figure}
    %     \centering
    %     \includegraphics[width=\linewidth]{media/network-structure.pdf}
    %     \caption{Network Structure \twz{(To be replaced)}}
    %     \label{fig:network-struecutre}
    % \end{figure}

\subsection{Equivariant Modeling}
    Let the concatenated proprioceptive features be denoted by $\mathbf{Z} \in \mathbb{R}^{d}$, and let $G = D_{3}$ represent the dihedral symmetry group of order six corresponding to the robot’s morphology. Each group element $g \in G = \{e, r, r^{2}, f, fr, fr^{2}\}$ denotes a discrete rotation or reflection that leaves the tensegrity structure invariant up to relabeling.
    
    For each transformation $g$, we define a permutation operator $\pi_{g}$ acting on the node and feature indices such that $\pi_{g} : \mathbf{Z} \mapsto \pi_{g}(\mathbf{Z})$. A shared message-passing function $F(\cdot)$ processes each transformed representation, after which the outputs are inverse-transformed and averaged:
    \begin{equation}
        \hat{\mathbf{y}} 
        = \frac{1}{|G|} 
          \sum_{g \in G} 
          \pi_{g}^{-1} 
          F\!\big(\pi_{g}(\mathbf{Z})\big).
        \label{eq:equivariance}
    \end{equation}
    This formulation ensures the equivariance property
    \begin{equation}
        f(\pi_{g}\mathbf{Z}) = \pi_{g} f(\mathbf{Z}), 
        \quad \forall g \in G,
    \end{equation}
    so that symmetry-related input configurations produce correspondingly permuted predictions. The model therefore, respects the physical indistinguishability among endcaps, rods, and tendons under the group transformations.

    The network can thus be interpreted as an ensemble of $|G|$ parallel branches, each operating in a different symmetry frame, as shown in Fig.~\ref{fig:framework}. 
    Each branch applies $\pi_{g}$, performs identical message passing with shared weights, and maps the output back to canonical frame via $\pi_{g}^{-1}$. 
    Averaging over all inverse-aligned outputs enforces group-level consistency and invariance to node labeling, acting as an implicit regularizer that reduces the hypothesis space without explicit data augmentation.

\subsection{HGNN Encoder}
    The HGNN encoder learns latent representations of all nodes by propagating information along the tensegrity structure. 
    Its objective is to integrate multi-modal proprioceptive signals---rigid-body motion from rods and tension feedback from tendons---into consistent feature embeddings for the six endcaps.

    \subsubsection{Message-passing formulation}
        For each edge $(i,j)\in\mathcal{E}$ with feature vector $E_{i,j}$, the transmitted message is:
        \begin{equation}
            m_{i\rightarrow j} = \phi_m(V_i, E_{i,j}),
        \end{equation}
        where $V_i$ denotes the current embedding of node $i$, and $\phi_m(\cdot)$ is a learnable message function.
        Node $j$ aggregates incoming messages from its neighborhood $\mathcal{N}(j)$ and updates its embedding as:
        \begin{equation}
            V_j' = \phi_u\!\left(V_j, \sum_{i\in\mathcal{N}(j)} m_{i\rightarrow j}\right),
            % = \phi_u\left( V_j, \sum_{i\in\mathcal{N}(j)} \phi_m\left( V_i, E_{i,j}\right) \right),
        \end{equation}
        % where $\phi_u(\cdot)$ is an update operator that fuses the aggregated information with the node’s previous state. 
        where $\phi_u(\cdot)$ fuses the aggregated information with the previous node state.
        Edge features $E_{i,j}$ encode connection type(rod–endcap or tendon–endcap).
        By stacking $K$ layers of this operation (eight in our implementation), the encoder expands its receptive field, allowing global coupling to emerge from local interactions.

    \subsubsection{Heterogeneous propagation}
        Because tensegrity structure includes multiple connection types, messages are computed separately for each edge category.
        Let $t \in \{\text{rod--endcap},\, \text{tendon--endcap}\}$ denote an edge type, and let $W_t^{(k)}$ be its transformation matrix at layer $k$.
        The update for node type $a$ is:
        \begin{equation}
            V_a^{(k+1)} = \sigma\!\left(
                \sum_{t \in \mathcal{T}(a)} 
                W_t^{(k)}\, M_t^{(k)} V_{\mathcal{N}_t(a)}^{(k)}
            \right),
        \end{equation}
        where $M_t^{(k)}(\cdot)$ aggregates features from neighbors connected by edge type $t$. $\mathcal{N}_t(a)$ denotes the corresponding neighborhood, and $\sigma(\cdot)$ is a nonlinear activation function.
        This formulation enables the network to treat rigid and elastic couplings with distinct parameters while aggregating their effects at shared endcap nodes.

        The overall HGNN encoder $\Theta_\mathrm{HGNN}(\cdot)$ is the composition of $K$ message-passing layers parameterized by $\phi_m^{(k)}$ and $\phi_u^{(k)}$:
            \begin{equation}
            V^{(K)} =
            {\Theta_\mathrm{HGNN}}(V^{(0)},E,\mathcal{G}) =
            \phi_u^{(K)}\phi_m^{(K)} \cdots 
            \phi_u^{(1)} \phi_m^{(1)} V^{(0)},
            \end{equation}
        producing the final node embeddings $V^{(K)}$ that encode contact information.

\subsection{Output and Loss Function}
    
    The HGNN decoder outputs maps the final node embeddings $V^{(K)}$ to contact logits through a node-wise decoder function $f_{dec}(\cdot)$ (implemented as a MLP):
    \begin{equation}
        \begin{aligned}
               \hat{c}_i &= f_{dec}(V_i^{(K)}) \\
        \hat{\mathbf{C}} &= [\hat{c}_0, \hat{c}_1, \hat{c}_2, \hat{c}_3, \hat{c}_4, \hat{c}_5]^\top \in \mathbb{R}^{6},
        \end{aligned}
    \end{equation}
    with each element representing the predicted contact logit for one endcap.
    Given the ground-truth binary labels 
    \begin{equation}
        \mathbf{C} = [c_0, c_1, c_2, c_3, c_4, c_5]^\top, \quad c_i \in \{0,1\},
    \end{equation}
    training minimizes the Binary Cross-Entropy with Logits loss (BCEWithLogitsLoss):
    \begin{equation}
        \mathcal{L} = - \frac{1}{6} \sum_{i=0}^{5}
    \left[ c_i \log \sigma(\hat{c}_i) + (1 - c_i)\log(1 - \sigma(\hat{c}_i)) \right],
    \end{equation}
    where $\sigma(\cdot)$ denotes the sigmoid activation function.  
    This formulation optimizes all six contact classifiers while maintaining differentiability and numerical stability by combining the sigmoid and logarithmic terms within a single operation. 
    
    At inference, binary contact decisions are obtained by thresholding the predicted probabilities:
    \begin{equation}
    \tilde{c}_i = \mathbb{I}[\sigma(\hat{c}_i) > 0.5],
    \end{equation}
    yielding a binary contact vector $\tilde{\mathbf{C}} \in \{0,1\}^{6}$ for the six endcaps.  
    These predicted contact states are subsequently incorporated into the downstream Invariant EKF state estimator as measurement updates.

\section{Experimental Results}
    % \subsection{Experimental Setup}
    We train and evaluate the proposed Sym-HGNN for contact estimation as described in the previous section.
    The model employs eight message-passing layers, a 100-frame temporal history window, and a hidden dimension of 128 for both the encoder and decoder MLPs.
    Training uses a binary cross-entropy loss with a learning rate of $3\times10^{-4}$, batch size of 2048, and 30 epochs. All experiments are conducted on an NVIDIA RTX~5070 Ti GPU, requiring approximately 57 minutes to train on the complete motion-primitive dataset.
    
    Two experiments are considered for evaluation: cross-motion generalization and morphological generalization across turning radii. We then further investigate how contact estimation impacts the performance of downstream pose estimation.
    
    % We evaluate whether incorporating symmetry-equivariant structures into a heterogeneous graph neural network improves contact estimation for a three-bar tensegrity robot. 
    % The study focuses on two main aspects: (i) \textbf{prediction performance and sample efficiency}, assessing whether the proposed Sym-HGNN achieves higher classification accuracy and F1-scores than the baseline algorithms, particularly under limited training data; and (ii) \textbf{generalization capability}, examining whether symmetry-equivariant message passing enhances transfer across different motion primitives and geometric variations, such as turning radius. 
 
\subsection{Datasets}
    We collected IMU, tendon-length, and ground-truth contact data at 100 Hz in MuJoCo simulation while the tensegrity robot locomotes on flat ground. 
    To ensure fidelity to the real-world system, Gaussian noise is added to all simulated sensor measurements, where the noise parameters are identified from data collected on the physical robot platform.
    Two types of simulation datasets were collected to evaluate the proposed models under controlled conditions.
    \begin{enumerate}
        \item Motion-primitive dataset: six locomotion behaviors—forward (F), backward (B), forward-right (FR), forward-left (FL), backward-left (BL), and backward-right (BR);
        \item Unseen-radii dataset: sequences with different turning radii to test generalization.
    \end{enumerate}

    \begin{figure}
        \centering
        \includegraphics[width=\linewidth]{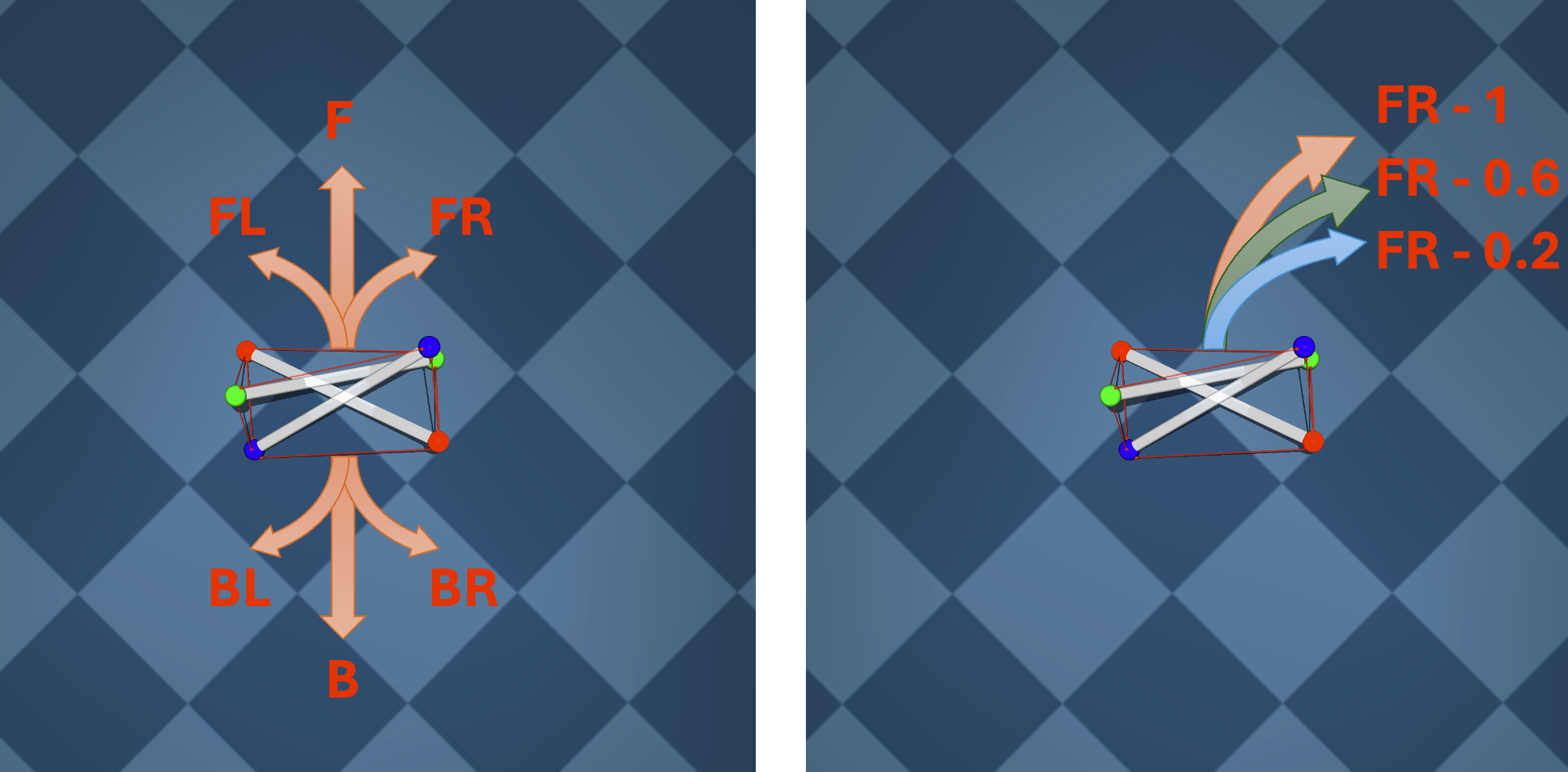}
        \caption{Simulation datasets with diverse motion primitives (left) and turning radii (right). The left plot illustrates six motion directions, while the right plot shows forward-right (FR) gaits with different turning radii (1.0, 0.6, and 0.2).}
        \label{fig:dataset}
    \end{figure}

\subsection{Models}
    Two baseline networks and our proposed network are evaluated in the following simulation contact experiments. 
    \begin{enumerate}
        \item A Convolutional Neural Network (CNN) baseline that processes concatenated IMU and tendon features through temporal convolutions;
        \item The Morphology-Informed HGNN (MI-HGNN), which encodes robot structure without explicit symmetry constraints; and 
        \item The Symmetry-Equivariant HGNN (Sym-HGNN), which integrates D$_3$-group operations into message passing and feature aggregation.
    \end{enumerate}
    
\subsection{Cross-Motion Generalization Experiments}
    We evaluate the model's generalization capability under limited motion diversity and limited training data. 
    Each model is trained on different subsets of six motion primitives: forward (F), backward (B), forward-left (FL), forward-right (FR), backward-left (BL), and backward-right (BR), and evaluated on a mixed test set that includes sequences from all six primitives. 
    This pipeline jointly examines both sample efficiency (performance with fewer training samples) and motion primitive diversity efficiency (performance when trained on fewer motion modes).
    
    Table~\ref{tab:sim_crossprimitive} summarizes the simulation results. 
    For each motion-primitive subset, the CNN baseline is trained exhaustively on all available samples within that subset, providing a fair upper bound for data utilization in conventional architectures. 
    In the all-primitives setting, we additionally trained CNN models on 30k and augmented datasets to match the reduced-data conditions of the GNNs; however, their performance remained substantially below that of the proposed models (accuracy $\approx$ 0.07–0.10, F1 $\approx$ 0.51–0.52), reflecting limited sample efficiency.
    By contrast, Sym-HGNN trained with only 10k–30k samples achieves 0.85–0.93 accuracy and 0.95–0.97 F1, comparable to or surpassing CNN trained on the full dataset.
    Across all primitive subsets, Sym-HGNN consistently outperforms MI-HGNN under equal data budgets and remains superior even to MI-HGNN trained with explicit data augmentation, indicating that the network effectively captures and exploits $D_3$ symmetry through equivariant message passing.
    Even under highly constrained motion diversity, such as training on only one or two primitives, Sym-HGNN maintains strong generalization, demonstrating that built-in symmetry constraints substantially enhance both sample efficiency and motion primitive diversity efficiency.

    % result_0.txt
    \begin{table}[t]
    \centering
    \setlength{\tabcolsep}{5pt}
    \renewcommand{\arraystretch}{1.1}
    \caption{
    \textbf{Cross-motion generalization in simulation.}
    Comparison of CNN, MI-HGNN, and Sym-HGNN trained on varying motion primitive subsets and sample sizes.
    Accuracy / F1 reported on the mixed test set.
    }
    \label{tab:sim_crossprimitive}
    \scriptsize
    \begin{tabularx}{\columnwidth}{
        @{}
        >{\arraybackslash}p{0.17\columnwidth}
        >{\centering\arraybackslash}p{0.23\columnwidth}
        >{\centering\arraybackslash}p{0.23\columnwidth}
        >{\centering\arraybackslash}p{0.23\columnwidth}}
    \toprule
    \textbf{Samples} & \textbf{CNN} & \textbf{MI-HGNN} & \textbf{Sym-HGNN} \\
    \midrule
    \rowcolor{gray!10}\multicolumn{4}{l}{\textbf{All primitives (F, B, FL, FR, BL, BR)}} \\
    137k (full) & 0.943 / 0.982 & 0.913 / 0.973 & 0.953 / 0.984 \\
    30k & 0.075 / 0.511 & 0.795 / 0.938 & \textbf{0.931 / 0.978} \\
    30k (aug) & 0.097 / 0.522 & 0.916 / 0.975 & — \\
    \midrule
    \rowcolor{gray!10}\multicolumn{4}{l}{\textbf{Three primitives (F, FR, FL)}} \\
    83k (full) & 0.836 / 0.923 & 0.810 / 0.936 & 0.942 / 0.980 \\
    15k & 0.078 / 0.513 & 0.607 / 0.850 & \textbf{0.891 / 0.964} \\
    15k (aug) & 0.060 / 0.502 & 0.821 / 0.947 & — \\
    30k &  0.072 / 0.510 & 0.718 / 0.899 & \textbf{0.915 / 0.971} \\
    30k (aug) & 0.076 / 0.501 & 0.888 / 0.966 & — \\
    \midrule
    \rowcolor{gray!10}\multicolumn{4}{l}{\textbf{Three primitives (F, FR, BR)}} \\
    56k (full) & 0.838 / 0.930 & 0.703 / 0.896 & 0.860 / 0.957 \\
    15k & 0.068 / 0.497 & 0.667 / 0.879 & \textbf{0.845 / 0.949} \\
    15k (aug) & 0.081 / 0.510 & 0.782 / 0.931 & — \\
    30k & 0.048 / 0.472 & 0.709 / 0.891 & \textbf{0.849 / 0.951} \\
    30k (aug) & 0.081 / 0.508 & 0.868 / 0.959 & — \\
    \midrule
    \rowcolor{gray!10}\multicolumn{4}{l}{\textbf{Two primitives (FL, FR)}} \\
    53k (full) & 0.814 / 0.911 & 0.790 / 0.925 & 0.902 / 0.967 \\
    20k & 0.078 / 0.507 & 0.662 / 0.871 & \textbf{0.876 / 0.959} \\
    20k (aug) & 0.068 / 0.500 & 0.870 / 0.957 & — \\
    \midrule
    \rowcolor{gray!10}\multicolumn{4}{l}{\textbf{Single primitive}} \\
    FL (40k full) & 0.807 / 0.903 & 0.750 / 0.912 & 0.885 / 0.962 \\
    FL (10k) & 0.103 / 0.486 & 0.479 / 0.790 & \textbf{0.852 / 0.949} \\
    FL (10k aug) & 0.078 / 0.520 & 0.784 / 0.928 & — \\
    % \cmidrule
    FR (14k full) & 0.491 / 0.744 & 0.355 / 0.703 & 0.610 / 0.848 \\
    FR (10k) & 0.161 / 0.328 & 0.347 / 0.692 & \textbf{0.626 / 0.843} \\
    FR (10k aug) & 0.062 / 0.505 & 0.575 / 0.836 & — \\
    \bottomrule
    \end{tabularx}
    \end{table}

\subsection{Turning Radii Generalization Experiments}
    A simulation dataset is collected with multiple turning radii. We define the turning ratio as the side-triangle length compared to the nominal turning gait. We collected \{FR, FL, BR, BL\} with turning ratios of \{1, 0.8, 0.6, 0.4\} and randomly split the data into training and validation sets with an 80/20 ratio. In a separate test set, we include \{FR, FL, BR, BL\} at a 0.2 turning ratio. 
    All models are trained on a 30k-sample subset from the entire training set and tested on the out-of-domain test set. Additionally, CNN, MI-HGNN, and Sym-HGNN are trained on the full dataset (143k samples) to assess the impact of data scale.
    
    % CNN, MI-HGNN and Sym-HGNN are trained on 30k samples covering several radii and tested on mixed-radius sequences whose specific radii are out of training domain. 

    Table~\ref{tab:sim_radii} presents the quantitative results. 
    Sym-HGNN achieves an average accuracy of 0.77 and an F1-score of 0.91, outperforming both CNN and MI-HGNN trained on the same 30k subset. 
    The CNN baseline shows poor sample efficiency and generalizability: reaching only 0.06 accuracy and 0.53 F1 when trained on 30k samples.
    These results demonstrate that symmetry-aware message passing siginificantly improves generalization to unseen data and offering robustness that cannot be attained through data scaling alone.
    
    \begin{table}[h]
    \centering
    \caption{Generalization across turning radius in simulation:\\ evaluation on unseen-radius sequence.
    % \twz{new test done, change wording in paper}
    % \mgj{Are you running experiment to add ours with more data? wasn't it 90 plus?} \wzt{We don't, but we can. The testing set is out-of-domain so it's not as good as 90\%+. }\mgj{We should otherwise it's not conclusive from the table where we stand with more data.} \twz{Ok, may need 5 hours to train it, will update results here later.}
    }
    \label{tab:sim_radii}
    \begin{tabular}{lccc}
    \toprule
    Model & Training Data $\downarrow$ & Avg. Accuracy $\uparrow$ & Avg. F1 $\uparrow$ \\
    \midrule
    CNN & \textbf{30k} & 0.06 & 0.53 \\
    CNN (full) & 143k & 0.80 & 0.93 \\
    \midrule
    MI-HGNN & \textbf{30k} & 0.44 & 0.77 \\
    MI-HGNN (full) & 143k & 0.70 & 0.89 \\
    \midrule
    \rowcolor{row_hightlight}
    \textbf{Sym-HGNN} & \textbf{30k} & \textbf{0.77} & \textbf{0.91} \\
    \rowcolor{row_hightlight}
    \textbf{Sym-HGNN (full)} & 143k & \textbf{0.85} & \textbf{0.94} \\
    \bottomrule
    \end{tabular}
    \end{table}
    
    % Across both tasks, Sym-HGNN consistently outperforms both CNN and MI-HGNN under identical data budgets and demonstrates improved robustness to both motion diversity and geometric variations. 
    % The CNN baseline, while achieving competitive results with full training data, exhibits poor sample efficiency and limited generalization under distribution shift.
    % These findings validate that embedding D$_3$ symmetry into the message-passing process provides a strong inductive bias for contact estimation in tensegrity structures, enabling both data-efficient learning and robust generalization.

\subsection{Ablation Study}
    % \begin{itemize}
    %     \item B1: InEKF with groundtruth contact label.
    %     \item B2: InEKF without contact aiding. (?)
    %     \item B3: InEKF with contact detected by CNN.
    %     % \item A1: Removal of symmetry averaging.
    %     % \item A2: Replacement of the 64-class head with six independent sigmoid heads.
    %     \item A3: Reduced temporal window ($L=25, 50, etc.$).
    %     % \item A4: IMU-only configuration (tendon features removed).
    %     \item A5: graph conv layer numbers
    % \end{itemize}

    Table~\ref{tab:ablation-study-of-network} summarizes the ablation results on network depth and temporal history length. Increasing the number of message-passing layers initially improves generalization, as deeper propagation enables richer coupling among rods, tendons, and endcaps. Performance peaks at eight layers, achieving the highest test F1-score of 0.9777 with moderate inference cost (34.8 ms). Further increasing depth to ten layers yields no significant gain and slightly degrades efficiency due to over-smoothing and higher computational load. Reducing the layer count to four or six leads to lower accuracy, indicating insufficient structural context aggregation.
    
    For temporal history, incorporating longer sequences enhances temporal consistency, with the best F1-score (0.9802) obtained using 200 frames of history. However, beyond 100 frames, accuracy gains are marginal, while memory usage increases. Overall, an 8-layer architecture with a 100-frame history offers the best trade-off between accuracy and runtime efficiency.

    % shorter version
    % Table~\ref{tab:ablation-study-of-network} shows the effect of network depth and history length on contact estimation. Performance improves with deeper message passing up to eight layers, which achieves the best F1-score (0.9777) and balanced efficiency (34.8 ms). Additional depth yields negligible gains but higher cost, while shallower models lose accuracy. Increasing the history window enhances temporal consistency, peaking at 200 frames (F1 = 0.9802), though improvements beyond 100 frames are minor. Overall, the 8-layer, 100-frame configuration provides the best trade-off between accuracy and runtime.
    
    \begin{table}[htbp]
        \centering
        \caption{Ablation study of network depth and history data length on contact estimation performance}
        \begin{tabular}{cccccc}
            \toprule
            \#Layers & \makecell{History\\Length} & \makecell{Train\\Acc.$\uparrow$} & \makecell{Test\\Acc.$\uparrow$} & \makecell{Test\\F1$\uparrow$} & \makecell{Avg. Inference\\Time$\downarrow$} \\
            \midrule
            10& 100 & 0.9931 & \textbf{0.9313} & 0.9776 & 40.83 ms\\ 
            \rowcolor{row_hightlight}
            8 & 100 & 0.9966 & 0.9306 & \textbf{0.9777} & 34.76 ms\\
            6 & 100 & \textbf{0.9970} & 0.9190 & 0.9737 & 28.28 ms\\
            4 & 100 & 0.9805 & 0.8612 & 0.9586 & \textbf{22.41} ms\\
            \midrule
            8 & 200 & \textbf{0.9980} & 0.9184 & \textbf{0.9802} & \textbf{34.33} ms\\
            \rowcolor{row_hightlight}
            8 & 100 & 0.9966 & \textbf{0.9306} & 0.9777 & 34.76 ms\\
            8 & 50  & 0.9917 & 0.8715 & 0.9472 & 34.63 ms\\
            8 & 25  & 0.9746 & 0.8190 & 0.9093 & 34.68 ms\\
            \bottomrule
        \end{tabular}
        \label{tab:ablation-study-of-network}
    \end{table}

\subsection{State Estimation with Learned Contacts}
    To evaluate the downstream impact of contact prediction, we integrate the learned contact estimates into the contact-aided Invariant Extended Kalman Filter (InEKF) for tensegrity robot pose estimation\cite{tong2025tensegrity, lin2023proprioceptive, hartley2020contact}. The test sequence involves straight-line rolling of the 3-bar tensegrity robot on flat terrain in MuJoCo simulation. 
    
\begin{figure}[htbp]
        \centering
        \includegraphics[width=\linewidth]{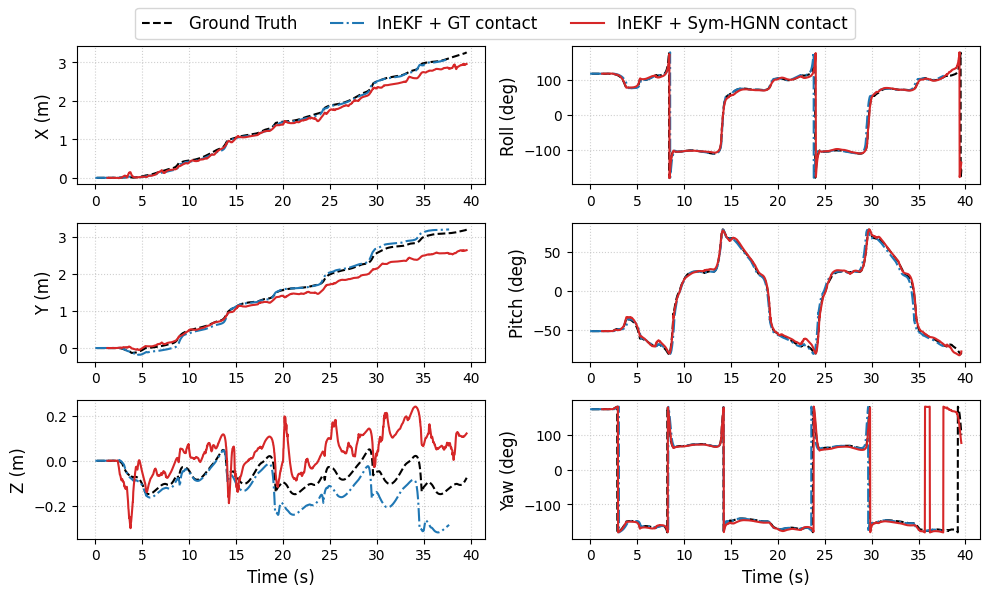}
        \caption{Estimated pose of the 3-bar tensegrity robot using the contact-aided Invariant Extended Kalman Filter (InEKF). The Sym-HGNN contact predictions provide observation updates that maintain consistency with ground-truth trajectories.}
        \label{fig:pose-xyz}
    \end{figure}

    As shown in Fig.~\ref{fig:pose-xyz}, the InEKF using HGNN-predicted contacts closely tracks the ground-truth trajectory across all translational and rotational states.  
    The estimated x–y positions and roll–pitch–yaw angles remain consistent with both the motion-capture groundtruth reference and the filter aided by ground-truth contacts, indicating that the learned contact events provide reliable update triggers for state correction.  
    Quantitatively, the final drift is 2.56\% for the InEKF with ground-truth contacts and 6.02\% when using Sym-HGNN predictions, demonstrating that the proposed contact estimation achieves comparable accuracy.
    However, minor deviations appear in vertical z displacement. 
    Potential sources of error contributing to these z-axis discrepancies include the unmodeled robot compliant deformation, contact impact on the IMU sensor, insufficient motion excitation along the z-axis compared to the x-y plane, and drift in estimated contact points.
    Despite these limitations, the overall drift remains low, demonstrating that robust, proprioceptively inferred contacts can effectively replace direct contact sensing for stable pose estimation in tensegrity robots.
    
    % \begin{figure}[htbp]
    %     \centering
    %     \includegraphics[width=\linewidth]{media/pose_xy_trajectory.png}
    %     \caption{Trajectories comparison on X-Y plane.}
    %     \label{fig:pose-xy}
    % \end{figure}

% \subsection{Real-world Experiments}
% TBD
% Unsuccessful contact experiments due to robot configuration sim2real gap

\section{Conclusion}
    This paper presented a symmetry-aware heterogeneous graph neural network (Sym-HGNN) for proprioceptive contact estimation in a three-bar tensegrity robot. The proposed model achieved strong sample efficiency and robust generalization without requiring dedicated contact sensors. Simulation results demonstrated that Sym-HGNN significantly outperforms both a CNN and MI-HGNN baselines across diverse locomotion primitives and unseen motions, achieving up to 15\% higher accuracy and 5\% higher F1-score with only 20\% of the training data. These contact predictions were then incorporated into a contact-aided Invariant Extended Kalman Filter to yield reliable state estimation performance compariable to the ground-truth contact results.

However, we also observed that the contact-prediction inference remains relatively slow, and the elevated false-positive rate impacts the accuracy of the state-estimation pipeline. 

Future work will include reducing inference latency and further lowering false-positive rates to enable real-time deployment on the physical tensegrity robot.
Additionally, we plan to fine-tune the model with a small set of real-world data, potentially collected via motion capture system or contact sensors, to address the sim-to-real gap and enhance the performance on real robot.
    
%%%%%%%%%%%%%%%%%%%%%%%%%%%%%%%%%%%%%%%%%%%%%%%%%%%%%%%%%%%%%%%%%%%%%%%%%%%%%%%%

% \section*{Acknowledgment}
% The authors would like to thank 

%%%%%%%%%%%%%%%%%%%%%%%%%%%%%%%%%%%%%%%%%%%%%%%%%%%%%%%%%%%%%%%%%%%%%%%%%%%%%%%%

\bibliographystyle{IEEEtran}
\bibliography{ref}

\end{document}